\documentclass[10pt,twocolumn,letterpaper]{article}

\usepackage[pagenumbers]{style} % To force page numbers, e.g. for an arXiv version

\usepackage[pagebackref,breaklinks,colorlinks]{hyperref}

\usepackage{multirow}
\usepackage{wrapfig}
\usepackage{colortbl}
\usepackage{makecell}
\usepackage{enumitem}
\usepackage{graphicx}
\usepackage{amsmath}
\usepackage{amssymb}
\usepackage{booktabs}

\usepackage{amsfonts}       % blackboard math symbols
\usepackage{nicefrac}       % compact symbols for 1/2, etc.
\usepackage{microtype}      % microtypography
\usepackage{xcolor}         % colors

\definecolor{Graylight}{gray}{0.9}
\usepackage{arydshln}
\newcommand{\tabincell}[2]{\begin{tabular}{@{}#1@{}}#2\end{tabular}}

\usepackage{algorithm}
\usepackage{algpseudocode}

\usepackage{appendix}
\usepackage{setspace}

\usepackage{listings}
\usepackage{wrapfig}
\definecolor{codeblue}{rgb}{0.25,0.5,0.5}
\definecolor{codekw}{rgb}{0.85, 0.18, 0.50}
\usepackage{subcaption}
\usepackage{wrapfig}

\usepackage[capitalize]{cleveref}
\crefname{section}{Sec.}{Secs.}
\Crefname{section}{Section}{Sections}
\Crefname{table}{Table}{Tables}
\crefname{table}{Tab.}{Tabs.}

\begin{document}

\title{Vision Transformer with Super Token Sampling}

\author{Huaibo Huang$^{1,2}$ ~~~ Xiaoqiang Zhou$^{1,4}$ ~~~ Jie Cao$^{1,2}$ ~~~  Ran He$^{1,2,3}$\footnotemark[1] ~~~ Tieniu Tan$^{1,2,4,5}$   \\
            {$^1$}MAIS\&CRIPAC, Institute of Automation, Chinese Academy of Sciences, China\\
            {$^2$}School of Artificial Intelligence, University of Chinese Academy of Sciences, China\\
            {$^3$}School of Information Science and Technology, ShanghaiTech University, China\\
            {$^4$}University of Science and Technology of China, Hefei, China \\
            {$^5$}Nanjing University, Nanjing, China \\
            {\tt\small \{huaibo.huang, jie.cao\}@cripac.ia.ac.cn, xq525@mail.ustc.edu.cn, \{rhe, tnt\}@nlpr.ia.ac.cn }
}

\maketitle

\let\thefootnote\relax
\footnotetext{* Ran He is the corresponding author.}

%%%%%%%%% ABSTRACT
\begin{abstract}

Vision transformer has achieved impressive performance for many vision tasks. However, it may suffer from high redundancy in capturing local features for shallow layers. Local self-attention or early-stage convolutions are thus utilized, which sacrifice the capacity to capture long-range dependency. A challenge then arises: can we access efficient and effective global context modeling at the early stages of a neural network? To address this issue, we draw inspiration from the design of superpixels, which reduces the number of image primitives in subsequent processing, and introduce super tokens into vision transformer. Super tokens attempt to provide a semantically meaningful tessellation of visual content, thus reducing the token number in self-attention as well as preserving global modeling.
Specifically, we propose a simple yet strong super token attention (STA) mechanism with three steps:
the first samples super tokens from visual tokens via sparse association learning, the second performs self-attention on super tokens, and the last maps them back to the original token space. STA decomposes vanilla global attention into multiplications of a sparse association map and a low-dimensional attention, leading to high efficiency in capturing global dependencies.
Based on STA, we develop a hierarchical vision transformer.
Extensive experiments demonstrate its strong performance on various vision tasks.
In particular, it achieves \textbf{86.4\%} top-1 accuracy on ImageNet-1K without any extra training data or label, \textbf{53.9} box AP and \textbf{46.8} mask AP on the COCO detection task, and \textbf{51.9} mIOU on the ADE20K semantic segmentation task. %Code will be released at \url{https://github.com/hhb072/STViT}.

%Based on STA, we develop a hierarchical vision transformer architecture. Extensive experiments demonstrate its strong performance on different vision tasks, including image classification, object detection, instance segmentation, and semantic segmentation.

%Vision transformer has achieved impressive performance for many vision tasks. However, it may suffer from high redundancy in capturing local features for shallow layers. Local self-attention or early-stage convolutions are thus utilized, which sacrifice the capacity to capture long-range dependency. A challenge then arises: can we access efficient and effective global context modeling at the early stages of a neural network? To address this issue, we draw inspiration from the design of superpixels, which reduces the number of image primitives in subsequent processing, and introduce super tokens into vision transformer. Super tokens attempt to provide a semantically meaningful tessellation of visual content, thus reducing the token number in self-attention as well as preserving global modeling. Specifically, we sample super tokens from visual tokens via a fast sampling algorithm, perform self-attention on super tokens, and finally map them back to the original token space. Based on super tokens, we develop a hierarchical vision transformer architecture. Extensive experiments demonstrate its strong performance on different vision tasks, including image classification, object detection, instance segmentation, and semantic segmentation.

\end{abstract}

%%%%%%%%% BODY TEXT
\section{Introduction}
\label{sec:intro}

Transformer~\cite{vaswani2017attention} has dominated the natural language processing field and shown excellent capability of capturing long-range dependency with self-attention.
Recent studies~\cite{dosovitskiy2020image,liu2021swin} have demonstrated that transformer can be successfully transplanted to vision scenarios.
Dosovitskiy et al.~\cite{dosovitskiy2020image} present the pioneering vision transformer (ViT), where self-attention performs global comparison among all visual tokens.
ViT and the following works~\cite{touvron2021training,touvron2021going} exhibit the strong capacity of learning global dependency in visual content, achieving impressive performance in many vision tasks \cite{carion2020end, zhu2020deformable,strudel2021segmenter, zheng2021rethinking, wang2021end}.
%, such as image classification~\cite{touvron2021training,touvron2021going},  object detection \cite{carion2020end, zhu2020deformable}, and semantic segmentation \cite{strudel2021segmenter, zheng2021rethinking, wang2021end}.
%Nevertheless, there remain vital challenges for ViTs, such as the heavy computational cost for high-resolution inputs and the inefficient encoding for local features in the shallow layers~\cite{raghu2021vision,li2022uniformer}.
Nevertheless, the computational complexity of self-attention is quadratic to the number of tokens, resulting in huge computational cost for high-resolution vision tasks, e.g., object detection and segmentation.

\begin{figure}[t]
\captionsetup{font=small}%
\scriptsize
\begin{center}
\includegraphics[width=0.4\textwidth]{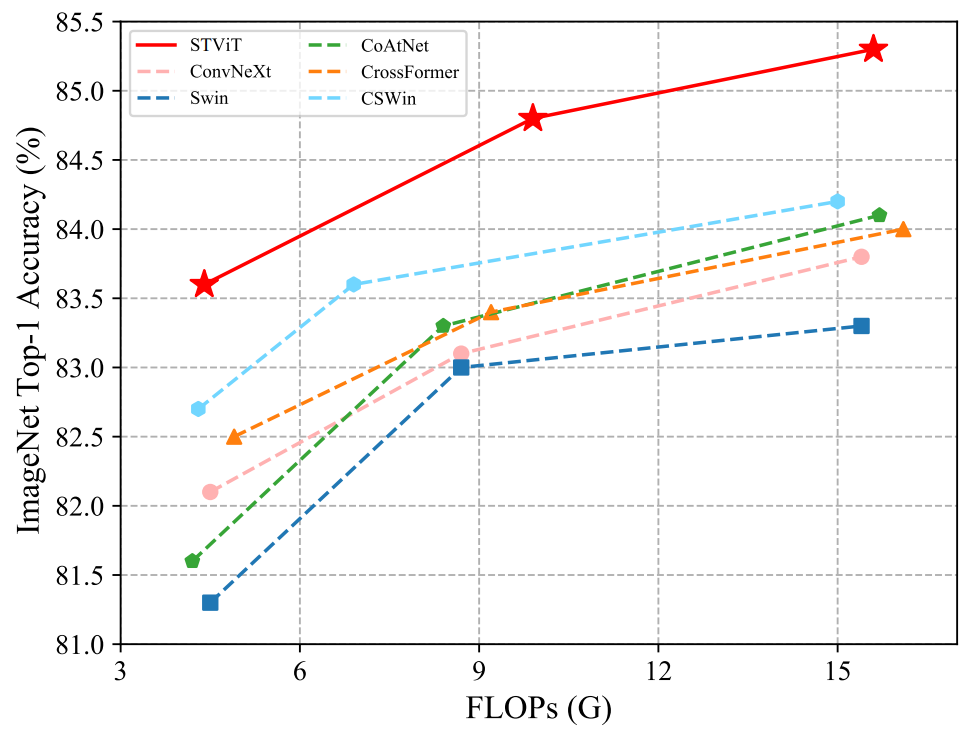}\label{fig:param_psnr}%
\hspace{-30mm}\resizebox{.28\columnwidth}{!}{
\begin{tabular}[b]{lcc}
	\footnotesize
	Model & \#Params & Top1 Acc.\\
	\hline
	%ConvNext-T~\cite{liu2022convnet} & 28M & 82.1 \\
%	Swin-T~\cite{liu2021swin} & 29M & 81.3  \\
%	CoAtNet-0~\cite{dai2021coatnet} & 25M & 81.6  \\
%    CrossFormer-S~\cite{wang2021crossformer} & 31M & 82.5 \\
%    CSWin-T~\cite{dong2021cswin} & 23M & 82.7  \\
%    \textbf{STViT-S (Ours)} & \textbf{25M} & \textbf{83.6} \\
%	\hline
    ConvNext-S~\cite{liu2022convnet} & 50M & 83.1 \\
	Swin-S~\cite{liu2021swin} & 50M & 83.0  \\
	CoAtNet-1~\cite{dai2021coatnet} & 42M & 83.3  \\
    CrossFormer-B~\cite{wang2021crossformer} & 52M & 83.4 \\
    CSWin-S~\cite{dong2021cswin} & 35M & 83.6  \\
    \textbf{STViT-B (Ours)} & \textbf{52M} & \textbf{84.8} \\
    \hline
	ConvNext-B~\cite{liu2022convnet} & 89M & 83.8 \\
	Swin-B~\cite{liu2021swin} & 88M & 83.3 \\
	CoAtNet-2~\cite{dai2021coatnet} & 75M & 84.1 \\
    CrossFormer-L~\cite{wang2021crossformer} & 92M & 84.0 \\
    CSWin-B~\cite{dong2021cswin} & 78M & 84.2 \\
    \textbf{STViT-L (Ours)} & \textbf{95M} & \textbf{85.3} \\
	\hline
	\multicolumn{3}{c}{\vspace{15mm}}
\end{tabular}
}
\vspace{-0.3cm}
%\caption{\#Params vs. Top-1 accuracy on ImageNet.}
\caption{FLOPs vs. Accuracy on $224^2$ ImageNet-1K images.}
\label{fig:flops_psnr}
\end{center}
\vspace{-1.2cm}
\end{figure}

\begin{figure*}[t]
\centering
\begin{subfigure}{0.9\linewidth}
	\centering
    \normalsize
	\includegraphics[width=0.95\linewidth]{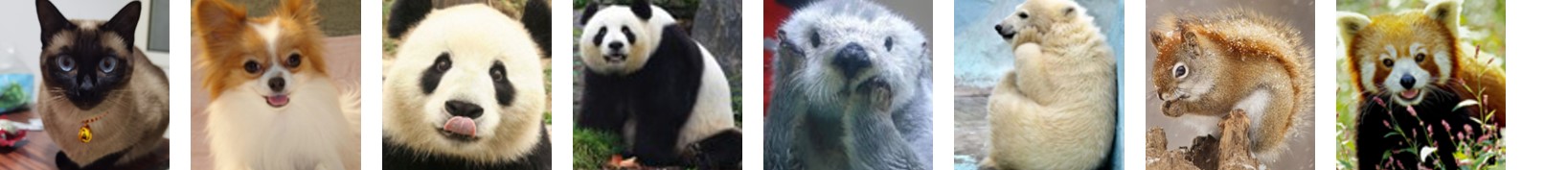}
	\caption{Input Images.}
	\label{fig:attn:input}%文中引用该图片代号
\end{subfigure}
\begin{subfigure}{0.9\linewidth}
	\centering
	\includegraphics[width=0.95\linewidth]{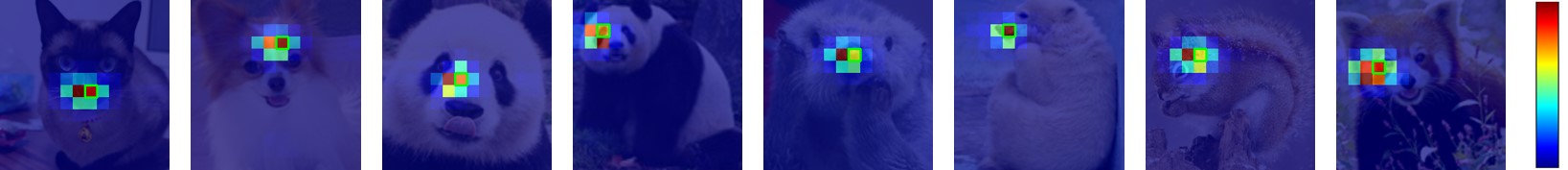}
	\caption{Attention maps from the third layer of  DeiT-S~\cite{touvron2021training}.}
	\label{fig:attn:deit}%文中引用该图片代号
\end{subfigure}
\begin{subfigure}{0.9\linewidth}
	\centering
	\includegraphics[width=0.95\linewidth]{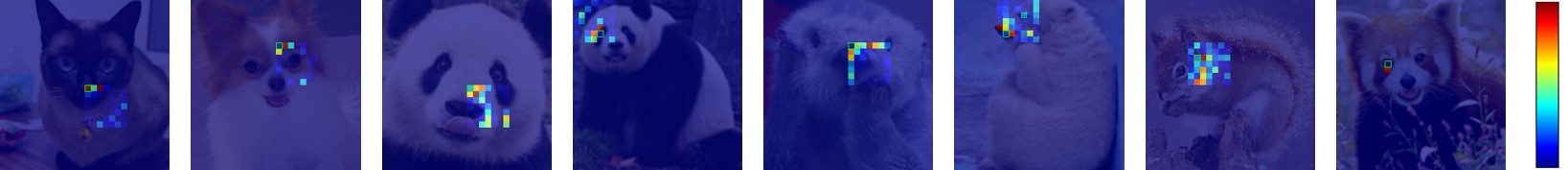}
	\caption{Attention maps from the third layer of  Swin-T~\cite{liu2021swin}.}
	\label{fig:attn:swin}%文中引用该图片代号
\end{subfigure}
\begin{subfigure}{0.9\linewidth}
	\centering
	\includegraphics[width=0.95\linewidth]{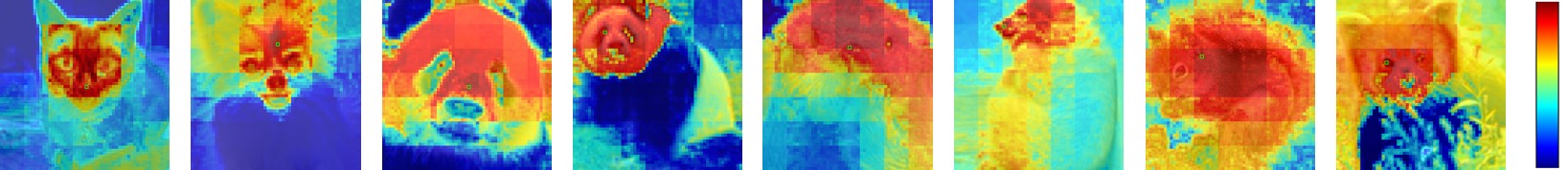}
	\caption{Attention maps from the third layer of STViT-S (Ours).}
	\label{fig:attn:STViT}%文中引用该图片代号
\end{subfigure}\vspace{-0.01\textwidth}
\caption{Visualization of early-stage attention maps for different vision transformers. For global attention in DeiT~\cite{touvron2021training} and local attention in Swin~\cite{liu2021swin}, only a few neighboring tokens (filled with red color) work for an anchor token (green box), resulting in local representations with high redundancy.
Compared with such ViTs, our method can learn global representations even for shallow layers.
}\vspace{-0.6cm}
\label{fig:attention}
\end{figure*}

%Specifically, self-attention has a quadratic computational complexity towards the number of tokens and the computational cost would increase to intolerably expensive for high-resolution vision tasks, e.g., object detection and segmentation.
Recent studies~\cite{raghu2021vision,li2022uniformer} observe that ViTs tend to capture local features at shallow layers with high redundancy. Specifically, as shown in Fig.~\ref{fig:attention}(b), given an anchor token, shallow-layer global attention concentrates on a few adjacent tokens (filled with red color) whereas neglects most of the tokens of far distance. Thus, global comparisons among all the tokens result in huge unnecessary computation cost in capturing such local correlations.
In order to reduce computation costs, Swin Transformer~\cite{liu2021swin} adopts window-based local attention to restrict attention to local regions. For local attention, as shown in Fig.~\ref{fig:attention}(c), redundancy is reduced but still exists in the shallow layers, where only a few nearby tokens obtain high weights.
In another way, Uniformer~\cite{li2022uniformer} utilizes  convolutions in the shallow layers and effectively reduces the computation redundancy for local features.
Nevertheless, both the local attention~\cite{liu2021swin} and early-stage convolution~\cite{li2022uniformer} schemes sacrifice the capacity of capturing global dependency that is crucial for transformer. A challenge then arises: can we access efficient and effective global representations at the early stages of a neural network?

To address this problem, inspired by the idea of superpixels~\cite{jampani2018superpixel}, we present super token attention to learn efficient global representations in vision transformer, especially for the shallow layers.
As an over-segmentation of image, superpixels perceptually group similar pixels together, reducing the number of image primitives for subsequent processing.
We borrow the idea of superpixels from the pixel space to the token space and assume super tokens as a compact representation of visual content.
We propose a simple yet strong super token attention (STA) mechanism with three steps.
Firstly, we apply a fast sampling algorithm to predict super tokens via learning sparse associations between tokens and super tokens. Then we perform self-attention in the super token space to capture long-range dependency among super tokens.
Compared to self-attention in the token space, such self-attention can reduce computational complexity significantly meanwhile learn global contextual information thanks to the representational and computational efficiency of super tokens.
At last, we map the super tokens back to the original token space by using the learned associations in the first step.
As shown in Fig.~\ref{fig:attention}(d), the presented super token attention can learn global representations even in the shallow layers. For example, given an anchor token, like the green box in the Siamese cat's nose, most of the related tokens (i.e., those in the cat face) contribute to the representation learning.

%Specifically, we apply a fast sampling algorithm to predict super tokens via aggregating semantically similar tokens together. We then perform self-attention in the super token space to capture long-range dependency among super tokens.
%Compared to self-attention in the token space, such self-attention can reduce computational complexity significantly meanwhile learn global contextual information thanks to the representational and computational efficiency of super tokens.
%Finally, we map the super tokens back to the original token space by using the token and super token associations learned in the sampling algorithm.
%As shown in Fig.~\ref{fig:attention}(d), the presented super token attention can learn global representations even in the shallow layers. For example, given an anchor token, like the green box in the nose of the Siamese cat, most of the tokens in the cat face would contribute to the representation learning.

Based on the super token attention mechanism, we present a general vision backbone named Super Token Vision Transformer (STViT) in this paper.
As shown in Fig.~\ref{fig:network}, it is designed as a hierarchical ViT hybrid with convolutional layers.
The convolutional layers  are adopted to compensate for the capacity of capturing local features.
In each stage, we use a stack of super token transformer (STT) blocks for efficient and effective representation learning. Specifically, the STT block consists of three key modules, i.e., Convolutional Position Embedding (CPE), Super Token Attention (STA), and Convolutional Feed-Forward Network (ConvFFN). The presented STA can efficiently learn global representations, especially for shallow layers. The CPE and ConvFFN with depth-wise convolutions can enhance the representative capacity of local features with a low computation cost.

Extensive experiments demonstrate the superiority of STViT on a broad range of vision tasks, including image classification, object detection, instance segmentation and semantic segmentation.
For example, without any extra training data, our large model STViT-L achieves \textbf{86.4\%} top-1 accuracy on ImageNet-1K image classification. Our base model STViT-B achieves \textbf{53.9} box AP and \textbf{46.8} mask AP on the COCO detection task, \textbf{51.9} mIOU on the ADE20K semantic segmentation task, surpassing the Swin Transformer \cite{liu2021swin} counterpart by  \textbf{+2.1}, \textbf{+2.1} and \textbf{+2.4}, respectively.

\section{Related Works}
\label{sec:relat}

\paragraph{Vision Transformers.} As the Transformer network achieves tremendous improvements in many NLP tasks~\cite{vaswani2017attention, devlin2018bert}, ViT~\cite{dosovitskiy2020image} is the pioneering work that adapts Transformer architecture to vision tasks and gets promising results. After that, given the great ability on long-range dependency modeling, researchers begin to pay more attention on designing general vision Transformer backbones~\cite{lee2022mpvit, yang2021focal}. Among them, many novel hierarchical architectures~\cite{liu2021swin, wang2021pyramid}, self-attention variants~\cite{ren2022shunted, guo2022cmt, xia2022vision, li2022mvitv2, yuan2021tokens} and positional encodings~\cite{vaswani2017attention, shaw2018self, chu2021conditional, dong2021cswin} are developed to accommodate the characteristics of vision tasks.

Many efficient self-attention mechanisms are introduced to alleviate the great computational costs. One way is restricting the attention region to spatially neighbor tokens, such as Swin Transformer~\cite{liu2021swin}, CSWin~\cite{dong2021cswin} and criss-cross attention~\cite{huang2019ccnet}. Besides, PVT~\cite{wang2021pyramid} designs a pyramid architecture with downsampled key and value tokens. GG Transformer~\cite{yu2021glance} and CrossFormer~\cite{wang2021crossformer} resort to dilated token sampling~\cite{zhao2021improved}.
%Moreover, clustering-based transformers~\cite{kitaev2020reformer,zeng2022not} are proposed to improve the efficiency.
Reformer~\cite{kitaev2020reformer} distributes tokens to buckets by hashing functions and perform dense attention within each bucket.
ACT~\cite{zheng2020end} and
TCFormer~\cite{zeng2022not} take the merged tokens as queries and the original tokens as keys and values to reduce the computation complexity.
Orthogonal Transformer~\cite{huang2022orthogonal} computes self-attention in the orthogonal space to learn better global and local representations.
Different from these works, the presented STA conducts sparse mapping between tokens and super tokens via soft associations and perform self-attention in the super token space. %It decomposes  global attention into multiplications of sparse and small matrices to learn efficient global representations.

%Unlike existing works, we sample super tokens by grouping similar tokens together and directly perform self-attention in the low-resolution super token space. Our method can not only reduce the computation burden of self-attention but also achieve effective global representations even in the shallow layers.

%However, these token reduction strategies are based on grid designment and neglect the contextual information in the image content. In this paper, we propose an efficient and perceptual token sampling mechanism, which is an extension of the classical superpixel~\cite{ren2003learning} in computer vision. Unlike existing methods, our method

\paragraph{Superpixel Algorithms.}
Traditional superpixel algorithms can be divided into two categories, i.e., graph-based and clustering-based methods. The graph-based methods view image pixels as graph nodes and partition nodes by the edge connectivity between adjacent pixels~\cite{ren2003learning, felzenszwalb2004efficient, liu2011entropy}. The clustering-based methods leverage traditional clustering techniques to construct superpixels, such as $k$-means clustering on different feature representation~\cite{achanta2012slic, liu2016manifold}. With the prosperity of deep learning in recent years, more and more deep clustering approaches~\cite{yeo2017superpixel, achanta2017superpixels, yang2020superpixel, cai2021revisiting} are proposed to exploit deep features and improve the clustering efficiency. SEAL~\cite{tu2018learning} proposes to learn the deep feature with a traditional superpixel algorithm. SSN~\cite{jampani2018superpixel} develops an end-to-end differentiable superpixel segmentation framework. Our sampling algorithm is mostly motivated by SSN~\cite{jampani2018superpixel} but we adopt a more attention-like manner with sparse sampling.

\section{Method}
\label{sec:metho}

\begin{figure*}[t]
\centering
\includegraphics[width=0.9\linewidth]{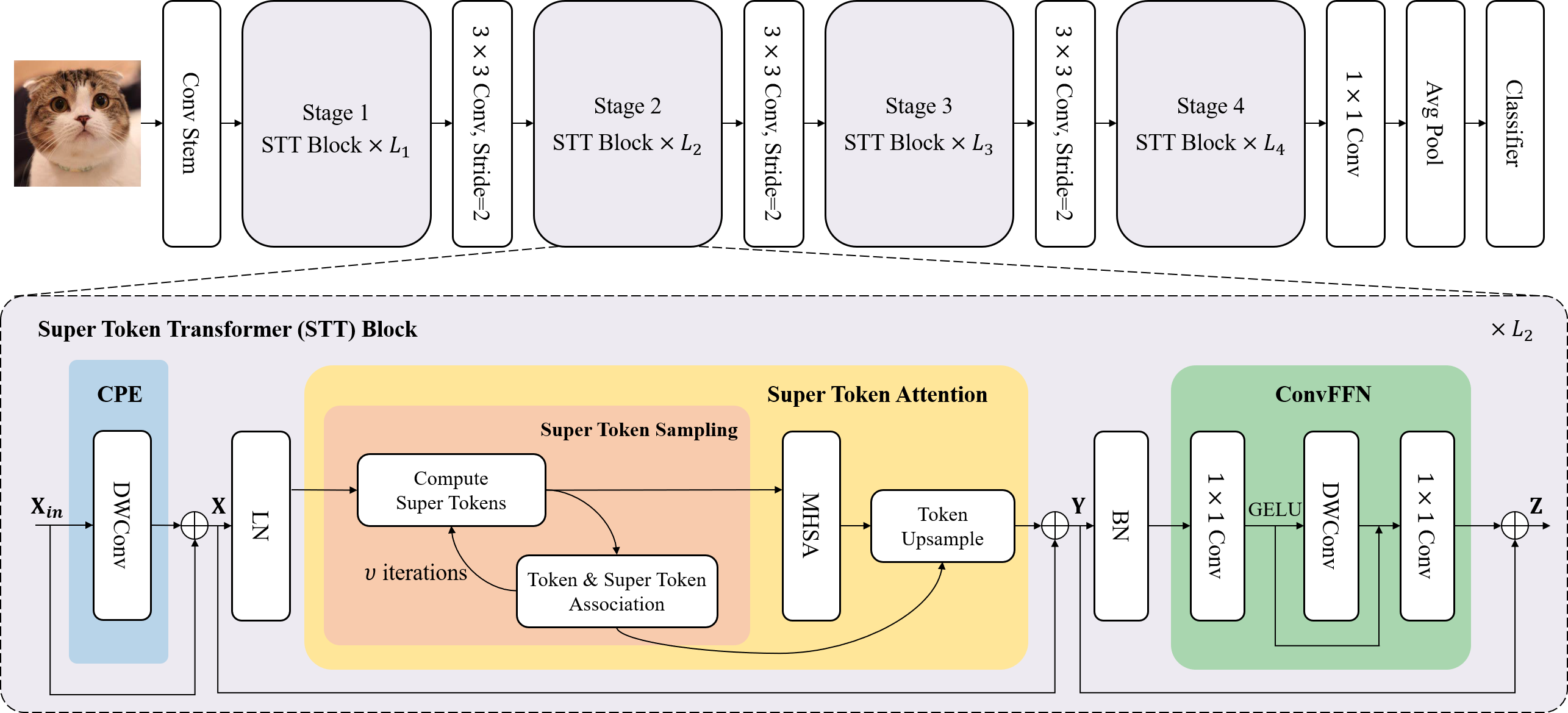}
\caption{The architecture of Super Token Vision Transformer (STViT).
}
\label{fig:network}
\end{figure*}

\subsection{Overall Architecture}
\label{sec:metho:archi}

An overview of the  Super Token Vision Transformer (STViT) architecture is illustrated in Fig.~\ref{fig:network}.
Given an input image, we firstly feed it into a stem consisting of four $3\times 3$ convolutions with stride 2, 1, 2, 1, respectively.
Compared with the vanilla non-overlapping tokenization~\cite{vaswani2017attention,liu2021swin}, the convolution stem can extract better local representations and is widely applied in recent ViTs~\cite{guo2022cmt,li2022uniformer}.
Then the tokens go through four stages of stacked Super Token Transformer (STT) blocks for hierarchical representation extraction. The $3\times 3$ convolutions with stride 2 are used between the stages to reduce the token number. Finally, the $1\times 1$ convolution projection~\cite{guo2022cmt}, global average pooling and fully connected layer are used to output the predictions.

A STT block contains three key modules: Convolutional Position Embedding (CPE), Super Token Attention (STA) and Convolutional Feed-Forward-Network (ConvFFN):
\begin{align}\label{eq:stt}
  X &= {\rm CPE}(X_{in}) + X_{in}, \\
  Y &= {\rm STA}({\rm LN}(X)) + X, \\
  Z &= {\rm ConvFFN}({\rm BN}(Y)) + Y.
\end{align}

Given the input token tensor $X_{in}\in \mathbb{R}^{C\times H\times W}$, we firstly use CPE~\cite{chu2021conditional}, i.e., a $3\times 3$ depth-wise convolution, to add position information into all the tokens. Compared with the absolute positional encoding (APE) \cite{vaswani2017attention} and  relative positional encoding (RPE) \cite{liu2021swin, shaw2018self}, CPE  can learn absolute positions by zero padding, which is more flexible for arbitrary input resolutions.
Then, we use STA to extract global contextual representations by efficiently exploring and fully exploiting long-range dependencies. The details of STA are elaborated in the following subsection. Finally, we adopt a convolution FFN~\cite{huang2022orthogonal} to enhance local representations.  It consists of two $1\times 1$ convolutions, one $3\times 3$ depth-wise convolution and one non-linear function, i.e., GELU.
Note that the two depth-wise convolutions in CPE and ConvFFN can compensate the capacity of local correlation learning. Thereby, the combination of CPE, STA, and ConvFFN enables STT  to capture both local and global dependencies.

\subsection{Super Token Attention}
\label{sec:metho:sta}

As shown in Fig.~\ref{fig:network}, the Super Token Attention (STA) module consists of three processes, i.e., Super Token Sampling (STS), Multi-Head Self-Attention (MHSA), and Token Upsampling (TU). Specifically, we firstly aggregate tokens into super tokens by STS, then perform MHSA to model global dependencies in the super token space and finally map the super tokens back to visual tokens by TU.

\subsubsection{Super Token Sampling}

In the STS process, we adapt the soft k-means based superpixel algorithm in SSN~\cite{jampani2018superpixel} from the pixel space to the token space.
Given the visual tokens $X\in \mathbb{R}^{N\times C}$ (where $N=H\times W$ is the token number), each token $X_i\in \mathbb{R}^{1\times C}$ is assumed to belong to one of $m$ super tokens $S\in \mathbb{R}^{m\times C}$, making it necessary to compute the $X$-$S$  association map $Q\in \mathbb{R}^{N\times m}$. First, we sample initial super tokens $S^0$ by averaging tokens in regular grid regions. If the grid size is $h\times w$, then the number of super tokens is  $m=\frac{H}{h}*\frac{W}{w}$.
Then we run the sampling algorithm iteratively with the following two steps:

\textbf{Token \& Super Token Association.}
In SSN~\cite{jampani2018superpixel}, the pixel-superpixel association at iteration $t$ is computed as
\begin{align}\label{eq:ssn}
  Q^t_{ij} = e^{-\|X_i-S_j^{t-1}\|^2}.
\end{align}
Different from SSN~\cite{jampani2018superpixel}, we apply a more attention-like manner to compute the association map $Q^t$, defined as
\begin{align}\label{eq:asso}
  Q^t = {\rm Softmax} (\frac{X{S^{t-1}}^{\rm T}}{\sqrt{d}}),
\end{align}
where $d$ is the channel number $C$.

\begin{figure}[t]
\centering
\includegraphics[width=0.9\linewidth]{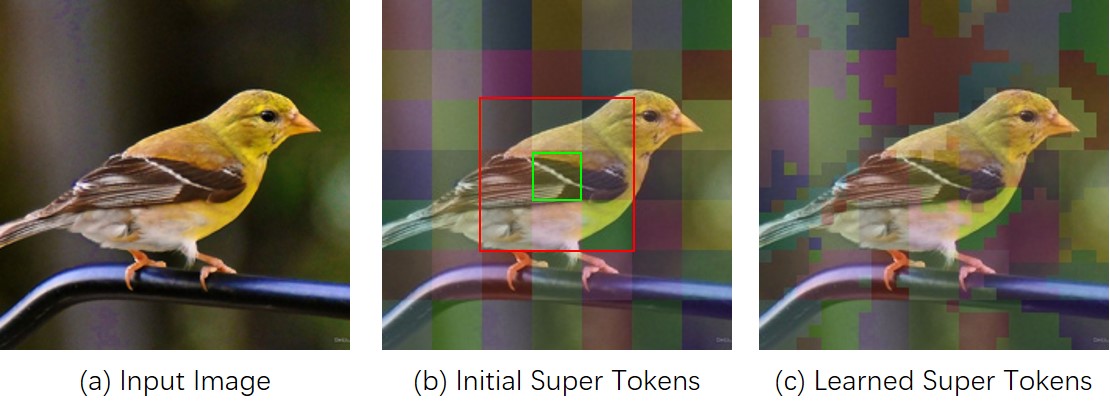}
\caption{Visualization of super tokens from initial grid to learned ones. Only the surrounding super tokens in the red box are used to compute the associations for every token in the green box.
}
\label{fig:stoken}
\end{figure}

\textbf{Super Token Update.} The super tokens are updated as the weighted sum of tokens, defined as
\begin{align}\label{eq:stoken}
  S = ({{\hat{Q}}^{t}})^{\rm T}X,
\end{align}
where $\hat{Q}^t$ is the column-normalized $Q^t$.
The computational complexity of the above sampling algorithm is
\begin{align}\label{eq:comp:sts0}
  \Omega ({\rm STS}) =  2 \upsilon mNC,
\end{align}
where $\upsilon$ is the number of iterations. It is time-consuming even for a small number of super tokens $m$.
To speed up the sampling process, following SSN~\cite{jampani2018superpixel}, we constrain the association computations from each token to only $9$ surrounding super tokens as shown in Fig.~\ref{fig:stoken}. For each token in the green box, only the super tokens in the red box are used to compute the association.  Moreover, we only  update the super tokens once with $\upsilon = 1$. Consequently, the complexity is significantly reduced to
\begin{align}\label{eq:comp:sts1}
  \Omega ({\rm STS}) =  19NC,
\end{align}
where the complexities for obtaining initial super-tokens, computing sparse associations and updating  super tokens are $NC$, $9NC$ and $9NC$, respectively. We provide the details of the sparse computation of STS in Appendix A.

\subsubsection{Self-Attention for Super Tokens}

Since super tokens are compact representations of visual content, applying self-attention to them can focus more on global contextual dependencies rather than local features.
%Since super tokens are compact representations that group local similar tokens together,  self-attention for them can focus more on capturing non-local contextual dependencies rather than local relationships, leading to better global representations.
We apply the standard self-attention to the sampled super tokens $S\in \mathbb{R}^{m\times C}$, defined as
\begin{align}
  {\rm Attn} (S) &= {\rm Softmax}(\frac{\mathbf{q}(S)\mathbf{k}^{\rm T}(S)}{\sqrt{d}}) \mathbf{v}(S) \\
          \label{eq:mhsa}       &= \mathbf{A}(S) \mathbf{v}(S)
\end{align}
where $\mathbf{A}(S)={\rm Softmax}(\frac{\mathbf{q}(S)\mathbf{k}^{\rm T}(S)}{\sqrt{d}})\in \mathbb{R}^{m\times m}$ is the attention map, $\mathbf{q}(S)=SW_q$, $\mathbf{k}(S)=SW_k$ and $\mathbf{v}(S)=SW_v$ are linear functions with parameters $W_q$, $W_k$ and $W_v$, respectively. We omit the multi-head setting for clarity.

\subsubsection{Token Upsampling}

Although super tokens can capture better global representations with self-attention, they have lost most of local details in the sampling process. Therefore, instead of directly using them as the input to the subsequent layers, we map them back into visual tokens and add them to the original tokens $X$.
We use the  association map $Q$ (we omit the iteration index since we only adopt one iteration) to upsample tokens from the super tokens $S$, which can be defined as
\begin{align}\label{eq:tu}
  {\rm TU}({\rm Attn}(S)) =  Q {\rm Attn}(S).
\end{align}

\begin{table}[t]
 \centering
 \caption{Architecture variants of STViT. The FLOPs are measured at resolution $224\times 224$. }
 \label{tab:arch}
 \resizebox{0.48\textwidth}{!}{
\begin{tabular}{l|ccc|cc}
\toprule
Model & Blocks & Channels & Heads &  Params   & FLOPs \\
\cmidrule(r){1-6}
STViT-S & [3,5,9,3] &  [64,128,320,512] &  [1,2,5,8]  & 25M & 4.4G \\
STViT-B & [4,6,14,6] & [96,192,384,512] &  [2,3,6,8]  & 52M & 9.9G \\
STViT-L & [4,7,19,8] & [96,192,448,640] &  [2,3,7,10]  & 95M & 15.6G \\
\bottomrule
\end{tabular}
}
\end{table}

\subsubsection{Analysis of Complexity and Redundancy}

In the following we compare our STA with the standard global self-attention and analyze the reason for its strong capacity of learning efficient global representations.

\textbf{Global Self-Attention.} The definition of  the standard global self-attention (GSA) is rewritten as
\begin{align}\label{eq:gsa}
  {\rm GSA} (X) = \mathbf{A}(X) \mathbf{v}(X),
\end{align}
where $\mathbf{A}(X)= {\rm Softmax}(\frac{\mathbf{q}(X)\mathbf{k}^{\rm T}(X)}{\sqrt{d}}) \in \mathbb{R}^{N\times N}$ is the attention map for the input tokens.
The computational complexity of GSA is
\begin{align}\label{eq:comp:gsa}
  \Omega ({\rm GSA}) = 2 N^2 C + 4NC^2.
\end{align}

\textbf{Super Token Attention.}
Considering the Eq.~(\ref{eq:stoken}), Eq.~(\ref{eq:mhsa}) and Eq.~(\ref{eq:tu}), the overall process of STA is reformulated as
\begin{align}
  {\rm STA} (X) &= Q (\mathbf{A}(S) ( \hat{Q}^{\rm T} X)  W_v) \\
          \label{eq:sta:overall}      &= \tilde{\mathbf{A}}(X) \mathbf{v}(X),
\end{align}
where $\tilde{\mathbf{A}}(X)=Q\mathbf{A}(S)\hat{Q}^{\rm T} \in \mathbb{R}^{N\times N}$ is the corresponding attention map for the input tokens.
The computational complexity of STA is
\begin{align}\label{eq:comp:sta}
   \Omega ({\rm STA}) &= \Omega ({\rm STS}) +  \Omega ({\rm MHSA}) + \Omega ({\rm TU}) \\
                      &= 19 NC + (2m^2C + 4mC^2) + 9 NC \\
                      &= 2m^2 C +   4mC^2 + 28NC.
\end{align}
Obviously, given $m$ smaller than $N$,  STA has a much lower computational cost than the global attention.

\textbf{Redundancy Discussion.} Images have large local redundancy, e.g., visual content tends to be similar in a local region. As shown in Fig.~\ref{fig:attention}(b), for a certain anchor token in the shallow layer, global attention would highlight a few tokens in a local region, resulting in high redundancy in the comparisons among all the tokens. Compared with visual tokens, the presented super tokens tend to have distinct patterns and suppress local redundancy. For example, as shown in Fig.~\ref{fig:stoken}, there is a significant discrepancy with the super tokens at the bird head and surrounding regions.
Thus, global dependencies can be better captured by self-attention in the super token space, as illustrated in Fig.~\ref{fig:attention}(d).
Moreover, considering Eq.~(\ref{eq:sta:overall}), our STA can be viewed as a specific global attention that decomposes a computationally expensive $N\times N$ attention $\mathbf{A}(X)$ into multiplications of sparse and small matrices, i.e., the sparse association $Q$ and the small $m\times m$ attention $\mathbf{A}(S)$, with far lower computation cost.
Therefore, we can conclude that STA can learn effective global representations with high efficiency.

\begin{table*}[t]
\renewcommand\arraystretch{.7}
 \centering
 %\tablestyle{6pt}{0.95}
 \caption{Performance comparison on ImageNet-1K classification. The throughput is measured on a single V100 GPU with batch size 16.  %The input size is $224\times 224$, except for those with $\uparrow384$.
 %$^*$ denotes using Token Labeling~\cite{jiang2021all}.
 }
 \label{tab:comp:imagenet}
 %\setlength\tabcolsep{12pt}
% \resizebox{0.5\linewidth}{!}{
%\resizebox{\linewidth}{!}{
\setlength{\tabcolsep}{8pt}{
\begin{tabular}[t]{clccccc}
\toprule
Model Size & Model & \#Param & Flops & Throughput  & Test Size & Top-1 \\
\midrule
\multirow{16}{*}{\rotatebox{90}{\tabincell{c}{small model size \\ ($\sim$25M)}}} &ConvNeXt-T~\cite{liu2022convnet}  & 28M & 4.5G & 720 & 224 & 82.1   \\
&DeiT-S~\cite{touvron2021training} & 22M & 4.6G  & 922 & 224 & 79.9  \\
&PVT-S~\cite{wang2021pyramid} & 25M & 3.8G& 722  & 224 & 79.8  \\
%T2T-14 [65] & 22M & 5.2G  & 224 & 80.7  \\
&Swin-T~\cite{liu2021swin} & 29M & 4.5G & 712  & 224 & 81.3  \\
%CvT-13~\cite{wu2021cvt} & 20M & 4.5G  & 224 & 81.6  \\
&CoAtNet-0~\cite{dai2021coatnet} & 25M & 4.2G& 943  & 224 & 81.6  \\
&Focal-T~\cite{yang2021focal} & 29M & 4.9G & 323 & 224 & 82.2  \\
%&CrossFormer-S~\cite{wang2021crossformer} & 31M & 4.9G& 636  & 224 & 82.5  \\
%RegionViT-S [5] & 31M & 5.3G  & 224 & 82.6  \\
% MViTv2-T~\cite{li2022mvitv2} & 24M & 4.7G & 513 & 224  & 82.3  \\
&DAT-T~\cite{xia2022vision} & 29M & 4.6G & 575 & 224 & 82.0  \\
&CSwin-T~\cite{dong2021cswin} & 23M & 4.3G & 515 & 224  & 82.7  \\
%DaViT-T~\cite{ding2022davit} & 28M & 4.5G & 224  & 82.8  \\
&UniFormer-S  \cite{li2022uniformer} & 24M & 4.2G & 824  & 224 & 82.9  \\
% Shunted-S~\cite{ren2022shunted} & 22M & 4.9G  & 224 & 82.9  \\
&MPViT-S~\cite{lee2022mpvit} & 23M & 4.7G & 352 & 224 & 83.0  \\
&Ortho-S~\cite{huang2022orthogonal} & 24M & 4.5G & 435  & 224 &83.4 \\
&CMT-S~\cite{guo2022cmt} & 25M & 4.0G & 481 & 224 & 83.5 \\
&\cellcolor{Graylight}STViT-S  & \cellcolor{Graylight}25M & \cellcolor{Graylight}4.4G & \cellcolor{Graylight}564  & \cellcolor{Graylight}224 & \cellcolor{Graylight}\textbf{83.6}  \\
% \hdashline
\cmidrule(r){2-7}
&CvT-13~\cite{wu2021cvt} & 20M & 16.3G& - & 384 & 83.0  \\
&CaiT-xs24~\cite{touvron2021going}  & 27M & 19.3G& 86 & 384 & 83.8  \\
& CoAtNet-0~\cite{dai2021coatnet} & 25M & 13.4G & 262 & 384 & 83.9  \\
% Shunted-S~\cite{ren2022shunted} & 22M & -G & 384 & 84.3  \\
&CSwin-T~\cite{dong2021cswin} & 23M & 14.0G& 204 & 384 & 84.3  \\
& \cellcolor{Graylight}STViT-S & \cellcolor{Graylight}25M & \cellcolor{Graylight}14.1G & \cellcolor{Graylight}188 &  \cellcolor{Graylight}384 & \cellcolor{Graylight}\textbf{85.0} \\

% \cmidrule(r){1-5}
\midrule

\multirow{16}{*}{\rotatebox{90}{\tabincell{c}{medium model size \\ ($\sim$50M)}}} & ConvNeXt-S~\cite{liu2022convnet} &  50M &  8.7G& 404  & 224 & 83.1   \\
&PVT-L~\cite{wang2021pyramid} & 61M & 9.8G& 321  & 224 & 81.7  \\
%T2T-24 [65] & 64M & 13.2G   & 224 & 82.2  \\
&Swin-S~\cite{liu2021swin} & 50M & 8.7G & 406 & 224 & 83.0  \\
%CvT-21~\cite{liu2021swin} & 32M & 7.1G   & 224 & 82.5  \\
&CaiT-s24~\cite{touvron2021going} & 47M & 9.4G& 326 & 224  & 82.7  \\
&CoAtNet-1~\cite{dai2021coatnet} & 42M & 8.4G & 471 & 224 & 83.3  \\
&Focal-S~\cite{yang2021focal} & 51M & 9.1G& 191  & 224 & 83.5  \\
&CrossFormer-B~\cite{wang2021crossformer} & 52M & 9.2G& 377  & 224 & 83.4  \\
&CSwin-S~\cite{dong2021cswin} & 35M & 6.9G & 315 & 224 & 83.6  \\
% MViTv2-S~\cite{li2022mvitv2} & 35M & 7.0G  & 224 & 83.6  \\
&DAT-S~\cite{xia2022vision} & 50M & 9.0G& 312 & 224 & 83.7  \\
&UniFormer-B  \cite{li2022uniformer}  & 50M & 8.3G & 375 & 224 &  83.9   \\
&Ortho-B~\cite{huang2022orthogonal} & 50M & 8.6G & 286 & 224  & 84.0 \\
% Shunted-B~\cite{ren2022shunted} & 40M & 8.1G  & 224 & 84.0  \\
% DaViT-S~\cite{ding2022davit} & 50M & 8.8G  & 224 & 84.2  \\
&MViTv2-B~\cite{li2022mvitv2} & 52M & 10.2G& 250 & 224 & 84.4  \\
&CMT-B~\cite{guo2022cmt} & 46M & 9.3G& 259 & 256 & 84.5 \\
& \cellcolor{Graylight}STViT-B & \cellcolor{Graylight}52M & \cellcolor{Graylight}9.9G & \cellcolor{Graylight}286  & \cellcolor{Graylight}224 & \cellcolor{Graylight}\textbf{84.8} \\
% \hdashline
\cmidrule(r){2-7}
%CvT-21~\cite{wu2021cvt} & 32M &  25.0G  & 384 &  84.9  \\
&CaiT-s24~\cite{touvron2021going} & 47M &  32.2G & 60 & 384 &  84.3  \\
&CSwin-S~\cite{dong2021cswin} & 35M & 22.0G& 128 & 384 & 85.0  \\
&CoAtNet-1~\cite{dai2021coatnet} & 42M & 27.4G & 124 & 384 & 85.1  \\
% Shunted-B~\cite{ren2022shunted} & 40M & 27.2G & 384 & 85.5  \\
&MViTv2-B~\cite{li2022mvitv2} & 52M & 36.7G&- & 384 & 85.6  \\
&\cellcolor{Graylight}STViT-B &\cellcolor{Graylight}52M &\cellcolor{Graylight}31.5G & \cellcolor{Graylight}95 &\cellcolor{Graylight}384 &\cellcolor{Graylight}\textbf{86.0} \\
% \cmidrule(r){1-5}
\midrule

\multirow{16}{*}{\rotatebox{90}{\tabincell{c}{large model size \\ ($\sim$100M)}}} & ConvNeXt-B~\cite{liu2022convnet} & 89M & 15.4G& 252  & 224 & 83.8  \\
&DeiT-B~\cite{touvron2021training} &  86M &  17.5G& 298   & 224 &  81.8  \\
&Swin-B~\cite{liu2021swin} &  88M &  15.4G& 258   & 224 &  83.3  \\
&CaiT-s48~\cite{touvron2021going} &  90M &  18.6G&  162 & 224  &  83.5  \\
&Focal-B~\cite{yang2021focal} &  90M &  16.0G  & 136 & 224 &  83.8  \\
&CrossFormer-L~\cite{wang2021crossformer} &  92M &  16.1G & 246  & 224 &  84.0  \\
&DAT-B~\cite{xia2022vision} & 88M & 15.8G & 211 & 224 & 84.0  \\
&CoAtNet-2~\cite{dai2021coatnet} & 75M & 15.7G & 298 & 224 & 84.1  \\
&CSwin-B~\cite{dong2021cswin} & 78M & 15.0G & 216 & 224 & 84.2  \\
&Ortho-L ~\cite{huang2022orthogonal}& 88M & 15.4G & 180 & 224  & 84.2 \\
&MPViT-B~\cite{lee2022mpvit} & 75M & 16.4G & 185 & 224 & 84.3  \\
% Shunted-L~\cite{ren2022shunted} & 81M & 14.9G  & 224 & 84.6  \\
% DaViT-B~\cite{ding2022davit} & 88M & 15.5G  & 224 & 84.6  \\
&CMT-L~\cite{guo2022cmt} & 75M & 19.5G & - & 288 & 84.8 \\
&\cellcolor{Graylight}STViT-L & \cellcolor{Graylight}95M & \cellcolor{Graylight}15.6G & \cellcolor{Graylight}192  & \cellcolor{Graylight}224  & \cellcolor{Graylight}\textbf{85.3} \\
% \hdashline
\cmidrule(r){2-7}
&Swin-B~\cite{liu2021swin} &  88M &  47.0G& 85  & 384 &  84.2  \\
% DAT-B~\cite{xia2022vision} & 88M & 49.8G & 384 & 84.8  \\
&CaiT-s48~\cite{touvron2021going} &  90M & 63.8G& 30 & 384  &  85.1  \\
&CSwin-B~\cite{dong2021cswin} & 78M & 47.0G& 69 & 384 & 85.5  \\
&CoAtNet-2~\cite{dai2021coatnet} & 75M & 49.8G & 88 & 384 & 85.7  \\
% Shunted-L~\cite{ren2022shunted} & 81M & 48.4G & 384 & 85.8  \\
&\cellcolor{Graylight}STViT-L  & \cellcolor{Graylight}95M & \cellcolor{Graylight}49.7G & \cellcolor{Graylight}65 & \cellcolor{Graylight}384 & \cellcolor{Graylight}\textbf{86.4}   \\
% \cmidrule(r){1-5}
\bottomrule
\end{tabular}
}
%}}
\end{table*} 

\begin{table*}[tp]
	\centering
    \caption{Object detection and instance segmentation with Mask R-CNN on COCO val2017. The FLOPs are measured at resolution 800$\times $1280. All the models are pre-trained on ImageNet-1K.
    }\vspace{-0.3cm}
    \label{results_detection_mask}
    \resizebox{1\textwidth}{!}{
    \begin{tabular}{l|cc|ccc|ccc|ccc|ccc}
        \toprule
        \multirow{2}{*}{Backbone} & \#Param & FLOPs & \multicolumn{6}{c|}{Mask R-CNN 1$\times$ schedule} & \multicolumn{6}{c}{Mask R-CNN 3$\times$ + MS schedule}\\
        ~ & (M) & (G) & $AP^b$ & $AP^b_{50}$ & $AP^b_{75}$ & $AP^m$ & $AP^m_{50}$ & $AP^m_{75}$ & $AP^b$ & $AP^b_{50}$ & $AP^b_{75}$ & $AP^m$ & $AP^m_{50}$ & $AP^m_{75}$ \\
	   \cmidrule(r){1-15}
        Res50 \cite{resnet} & 44 & 260  & 38.0 & 58.6 & 41.4 & 34.4 & 55.1 & 36.7
        & 41.0 & 61.7 & 44.9 & 37.1 & 58.4 & 40.1 \\
        PVT-S \cite{wang2021pyramid} & 44 & 245  & 40.4 & 62.9 & 43.8 & 37.8 & 60.1 & 40.3
        & 43.0 & 65.3 & 46.9 & 39.9 & 62.5 & 42.8 \\
        %TwinsP-S \cite{twins} & 44 & 245  & 42.9 & 65.8 & 47.1 & 40.0 & 62.7 & 42.9
        %& 46.8 & 69.3 & 51.8 & 42.6 & 66.3 & 46.0  \\
        %Twins-S \cite{twins} & 44 & 228  & 43.4 & 66.0 & 47.3 & 40.3 & 63.2 & 43.4
        %& 46.8 & 69.2 & 51.2 & 42.6 & 66.3 & 45.8 \\
        Swin-T \cite{liu2021swin} & 48 & 264  & 42.2 & 64.6 & 46.2 & 39.1 & 61.6 & 42.0
        & 46.0 & 68.2 & 50.2 & 41.6 & 65.1 & 44.8 \\
        %ViL-S \cite{zhang2021multi} & 45 & 218  & 44.9 & 67.1 & 49.3 & 41.0 & 64.2 & 44.1
        %& 47.1 & 68.7 & 51.5 & 42.7 & 65.9 & 46.2 \\
        Focal-T \cite{yang2021focal} & 49 & 291 & 44.8 & 67.7&  49.2 & 41.0 & 64.7
        & 44.2 & 47.2 & 69.4 & 51.9 & 42.7 & 66.5 & 45.9 \\
        CMT-S~\cite{guo2022cmt} & 45 & 249 & 44.6 & 66.8 & 48.9 & 40.7 & 63.9 & 43.4 & 48.3 & 70.4 & 52.3 & 43.7 & 67.7 & 47.1\\
        UniFormer-S~\cite{li2022uniformer} & 41 & 269 & 45.6 & 68.1 & 49.7 & 41.6 & 64.8 & 45.0 & 48.2 & 70.4 & 52.5 & 43.4 &67.1 &47.0 \\
       % CSWin-T~\cite{dong2021cswin} & 42 & 279 & 46.7 & 68.6 & 51.3 & 42.2 & 65.6 & 45.4 \\
       % RegionViT-S~\cite{chen2022regionvit} & 51 & 183 & 44.2 & - & - & 40.8 & - & - & 47.6 & - & - & 43.4 & - & - \\
        %CrossFormer-S~\cite{wang2021crossformer} & 50 & 291 & 45.0 & 67.9 & 49.1 & 41.2 & 64.6 & 44.3 & 48.7 & 70.7 & 53.7 & 43.9 & 67.9 & 47.3 \\
        \rowcolor{Graylight} STViT-S & 44 & 252 & \textbf{47.6} & \textbf{70.0} & \textbf{52.3} & \textbf{43.1} & \textbf{66.8} & \textbf{46.5}
        & \textbf{49.2} & \textbf{70.8} & \textbf{54.4} & \textbf{44.2} & \textbf{68.0} & \textbf{47.7}  \\
       \midrule
       Res101 \cite{resnet} & 63 & 336  & 40.4 & 61.1 & 44.2 & 36.4 & 57.7 & 38.8
        & 42.8 & 63.2 & 47.1 & 38.5 & 60.1 & 41.3\\
       % X101-32 \cite{resnext} & 63 & 340  & 41.9 & 62.5 & 45.9 & 37.5 & 59.4 & 40.2
        % & 44.0 & 64.4 & 48.0 & 39.2 & 61.4 & 41.9 \\
        PVT-M \cite{wang2021pyramid} & 64 & 302  & 42.0 & 64.4 & 45.6 & 39.0 & 61.6 & 42.1
        & 44.2 & 66.0 & 48.2 & 40.5 & 63.1 & 43.5 \\
        %TwinsP-B \cite{twins} & 64 & 302  & 44.6 & 66.7 & 48.9 & 40.9 & 63.8 & 44.2
        %& 47.9 & 70.1 & 52.5 & 43.2 & 67.2 & 46.3 \\
        % Twins-B \cite{twins} & 76 & 340  & 45.2 & 67.6 & 49.3 & 41.5 & 64.5 & 44.8
        % & 48.0 & 69.5 & 52.7 & 43.0 & 66.8 & 46.6  \\
        Swin-S \cite{liu2021swin} & 69 & 354  & 44.8 & 66.6 & 48.9 & 40.9 & 63.4 & 44.2
        & 48.5 & 70.2 & 53.5 & 43.3 & 67.3 & 46.6 \\
        Focal-S \cite{yang2021focal} & 71 & 401 & 47.4 & 69.8 & 51.9 & 42.8 & 66.6 & 46.1 & 48.8 & 70.5 & 53.6 & 43.8 & 67.7 & 47.2 \\
        DAT-S~\cite{xia2022vision} & 69 & 378 & 47.1 & 69.9 & 51.5 &  42.5 &  66.7 &  45.4 & 49.0 & 70.9 & 53.8 & 44.0 & 68.0 & 47.5 \\
        UniFormer-B~\cite{li2022uniformer} & 69 & 399 & 47.4 & 69.7 & 52.1 & 43.1 & 66.0 & 46.5 & 50.3 & \textbf{72.7} & 55.3 & 44.8 & 69.0 & 48.3 \\
        %RegionViT-B~\cite{chen2022regionvit} & 93 & 307 & 45.4 & - & - & 41.6 & - & - & 48.3 & - & - & 43.5 & - & -  \\
        %CrossFormer-B~\cite{wang2021crossformer} & 72 & 398 & 47.1 & 69.9 & 52.0 & 42.7 & 66.5 & 46.1 & 49.8 & 71.6 & 54.6 & 44.5 & 68.8 & 47.9  \\
        \rowcolor{Graylight} STViT-B & 70 & 359 & \textbf{49.7} & \textbf{71.7} & \textbf{54.7} & \textbf{44.8} & \textbf{68.9} & \textbf{48.7}
        & \textbf{51.0} & {{72.3}} & \textbf{56.0} & \textbf{45.4} & \textbf{69.5} & \textbf{49.3} \\
         \midrule
         Swin-B  & 107 &496 & 46.9 &- & - & 42.3&-&-&  48.5&69.8&53.2 & 43.4&66.8&46.9 \\
         CSWin-B & 97 & 526 & 48.7&70.4&53.9 & 43.9&67.8&47.3& 50.8& 72.1& 55.8 & 44.9& 69.1& 48.3 \\
         \rowcolor{Graylight} STViT-L  & 114 & 470 & \textbf{50.8} & \textbf{72.5} & \textbf{56.3} & \textbf{45.5} & \textbf{69.7} & \textbf{49.1} & \textbf{51.7} & \textbf{73.0} & \textbf{56.9} & \textbf{45.9} & \textbf{70.4} & \textbf{49.9}\\
        \bottomrule
    \end{tabular}
    }
\end{table*}

%\subsection{Hybrid with Convolutions}
%\label{sec:metho:cpe}

\subsection{Implementation Details}

As shown in Table~\ref{tab:arch}, we build three STViT backbones with different settings of block number and channel number in each stage. The expansion ratios in ConvFFN  are set to 4.
For the first two stages, the sampling grid sizes (as shown in Fig.~\ref{fig:stoken}(b)) are set to $8\times 8$ and $4\times 4$, respectively. Therefore, the number of super tokens $m$ is 49 for input resolution $224\times 224$.
For the last two stages, the local redundancy is already low after the  preceding two stages of  representation learning. So we set the grid sizes to $1\times 1$ and directly use the tokens as super tokens without the STS and TU processes.
The iteration number in STS is set to 1 to reduce the sampling cost.
More details are described in the Appendix B.

\section{Experiments}
\label{sec:exper}

We conduct experiments on a broad range of vision tasks, including image classification on ImageNet-1K \cite{deng2009imagenet}, object detection and instance segmentation on COCO 2017 \cite{lin2014microsoft}, and semantic segmentation on ADE20K \cite{zhou2019semantic}.
We also conduct ablation studies to examine the role of each component.

\subsection{Image Classification}

\textbf{Settings.}
We train our models from scratch on the ImageNet-1K~\cite{deng2009imagenet} data.
For a fair comparison, we follow the same training strategy proposed in DeiT~\cite{touvron2021training} and adopt the default data augmentation and regularization.
All our models are trained from scratch for 300 epochs with the input size of $224\times 224$.
We employ the AdamW optimizer with a cosine decay learning rate scheduler and 5 epochs of linear warm-up.
The initial learning rate,  weight decay, and  batch-size are  0.001, 0.05, and 1024, respectively.
The maximum rates of increasing stochastic depth \cite{huang2016deep} are set to 0.1/0.4/0.6 for STViT-S/B/L. When fine-tuning our models on $384\times 384$ resolution, the learning rate, weight decay, batch-size and total epoch are set to 5e-6, 1e-8, 512, 30, respectively. More details are in Appendix C.

\textbf{Results.}
We compare our STViT against the state-of-the-art models in Table~\ref{tab:comp:imagenet}.
The comparison results clearly show that our STViT outperforms previous models under different settings in terms of FLOPs and model size.
Specifically, our small model STViT-S achieves  \textbf{83.6\%} Top-1 accuracy with only 4.4G FLOPs, surpassing Swin-T, CSwin-T and MPViT-S by \textbf{2.3\%}, \textbf{0.9\%} and \textbf{0.6\%}, respectively. It achieves the same accuracy as Focal-S  with half of parameters, half of FLOPs and triplet speed.
Our base Model STViT-B achieves \textbf{84.8\%} accuracy, not only surpassing the corresponding counterparts with medium  model size but also surpassing those with large model size.
As for 384 $\times$ 384 input size, our large model STViT-L achieves an accuracy of \textbf{86.4\%}, surpassing Swin-B and CSwin-B by \textbf{2.2\%} and \textbf{0.9\%}, respectively, and outperforming CaiT-s48 by \textbf{1.3\%} with 22\% fewer FLOPs and double speed.
%The comparisons against the SOTA methods demonstrate the powerful learning capacity of our STViT.

\subsection{Object Detection and Instance Segmentation}

\textbf{Settings.} Experiments on object detection and instance segmentation are conducted on COCO 2017 dataset \cite{lin2014microsoft}.
%, which contains 118K training images and 5K validation images.
Following \cite{liu2021swin}, we use our models as the backbone network, and take Mask-RCNN~\cite{he2017mask} and Cascaded Mask R-CNN~\cite{cai2018cascade} as the detection and segmentation heads. The backbones are pretrained on ImageNet-1K, and fine-tuned on the COCO training set with the AdamW optimizer. We adopt two common experimental settings: ``1 $\times$'' (12 training epochs) and ``3 $\times$ +MS'' (36 training epochs with multi-scale training).
The configurations follow the setting used in Swin Transformer \cite{liu2021swin} and are implemented with MMDetection \cite{chen2019mmdetection}.

\textbf{Results.} We report the comparison results on the object detection task and instance segmentation task in Table~\ref{results_detection_mask} and Table~\ref{results_detection_cascade}. The results show that our STViT variants outperform all the other vision backbones.
For Mask R-CNN framework, our STViT-S outperforms Uniformer-S by \textbf{+2.0} box AP, \textbf{+1.5} mask AP under the 1$\times$ schedule.
For Cascade  Mask R-CNN framework, our STViT-B not only outperforms other backbones with similar FLOPs but also achieves better performance than those models having more FLOPs, e.g., outperforming Swin-B by \textbf{+2.0} box AP, \textbf{+1.5} mask AP and outperforming DAT-B by \textbf{+0.9} box AP, \textbf{+1.0} mask AP.

\begin{table}[tp]
	\centering
    \caption{Object detection and instance segmentation with Cascade Mask R-CNN on COCO val2017.
    }\vspace{-0.3cm}
    \label{results_detection_cascade}
    \resizebox{0.49\textwidth}{!}{
    \begin{tabular}{l|cc|ccc|ccc}
        \toprule
        \multirow{2}{*}{Method} & \#Params & FLOPs & \multicolumn{6}{c}{ 3$\times$ + MS schedule}\\
        ~ & (M) & (G) & $AP^b$ & $AP^b_{50}$ & $AP^b_{75}$ & $AP^m$ & $AP^m_{50}$ & $AP^m_{75}$ \\
	   \cmidrule(r){1-9}

       % %Res50 \cite{resnet} & 82 & 739 & 46.3 & 64.3 & 50.5 & 40.1 & 61.7 & 43.4 \\
%	    DeiT \cite{touvron2021training} & 80 & 889 & 48.0 & 67.2 & 51.7 & 41.4 & 64.2 & 44.3 \\
%	    Swin-T \cite{liu2021swin} & 86 & 745  & 50.5 & 69.3 & 54.9 & 43.7 & 66.6 & 47.1 \\
%	    %Shuffle-T \cite{shuffleTR} & 86 & 746  & 50.8 & 69.6 & 55.1 & 44.1 & 66.9 & 48.0 \\
%	    Focal-T \cite{yang2021focal} & 87 & 770  & 51.5 & 70.6 & 55.9 & - & - & - \\
%        DAT-T~\cite{xia2022vision} & 86 & 750 & 51.3& 70.1 &55.8 & 44.5 &67.5 &48.1 \\
%        %UniFormer-S~\cite{li2022uniformer} & 79 & 747 & 52.1 & 71.1 & 56.6 & 45.2 & 68.3 & 48.9 \\
%        \rowcolor{Graylight} STViT-S & 82 & 730 & & & & & &  \\
%        \midrule

        X101-32 \cite{resnext} & 101 & 819  & 48.1 & 66.5 & 52.4 & 41.6 & 63.9 & 45.2 \\
        Swin-S \cite{liu2021swin} & 107 & 838  & 51.8 & 70.4 & 56.3 & 44.7 & 67.9 & 48.5 \\
        Shuffle-S \cite{huang2021shuffle} & 107 & 844  & 51.9 & 70.9 & 56.4 & 44.9 & 67.8 & 48.6 \\
        CSwin-S \cite{dong2021cswin} & 92 &820  &53.7 &72.2& 58.4 &46.4& 69.6& 50.6 \\
        DAT-S~\cite{xia2022vision}  & 107 & 857 & 52.7 &71.7& 57.2  & 45.5& 69.1 &49.3 \\
        %UniFormer-B~\cite{li2022uniformer} & 107 &878& 53.8 &72.8& 58.5 &46.4& 69.9& 50.4 \\
        \textcolor{gray!70}{Swin-B}  \cite{liu2021swin} &  \textcolor{gray!70}{145} &  \textcolor{gray!70}{972} & \textcolor{gray!70}{51.9}&   \textcolor{gray!70}{70.9}&  \textcolor{gray!70}{57.0} & \textcolor{gray!70}{45.3} & \textcolor{gray!70}{68.5}&  \textcolor{gray!70}{48.9} \\
        \textcolor{gray!70} {DAT-B}~\cite{xia2022vision}  & \textcolor{gray!70} {145} & \textcolor{gray!70} {1003} & \textcolor{gray!70} {53.0}& \textcolor{gray!70} {71.9}& \textcolor{gray!70} {57.6} & \textcolor{gray!70} {45.8}& \textcolor{gray!70} {69.3}& \textcolor{gray!70} {49.5} \\
        \rowcolor{Graylight} STViT-B & 108  & 837  & \textbf{53.9} & \textbf{72.7} & \textbf{58.5} & \textbf{46.8} & \textbf{70.4} & \textbf{50.8}  \\
        \bottomrule
    \end{tabular}
    }\vspace{-0.3cm}
\end{table} 

\begin{table}[tp]
	\centering
    \caption{Semantic segmentation with  Upernet on ADE20K. The FLOPs are measured at resolution 512$\times$2048.}\vspace{-0.3cm}
    \label{results_segmentation}

    \resizebox{0.45\textwidth}{!}{
    \begin{tabular}{l|cc|cc}
        \toprule
        \multirow{2}{*}{Backbone}  &
         \begin{tabular}[c]{@{}c@{}}\#Param\\ (M)\end{tabular} & \begin{tabular}[c]{@{}c@{}}FLOPs\\ (G)\end{tabular} & \begin{tabular}[c]{@{}c@{}}mIoU\\ (\%)\end{tabular} & \begin{tabular}[c]{@{}c@{}}MS mIoU\\ (\%)\end{tabular}   \\	
        \midrule
        Swin-T  \cite{liu2021swin} & 60 & 945 & 44.5 & 45.8 \\
        CSWin-T~\cite{dong2021cswin} & 60 & 959 & \textbf{49.3} & \textbf{50.7} \\
        UniFormer-S \cite{li2022uniformer} & 52& 1008 & 47.6 & 48.5 \\
        \rowcolor{Graylight} SViT-S  & 54 & 926 & 48.6 & 49.0 \\
      \midrule
        Res101 \cite{resnet} & 86  & 1029 & - & 44.9 \\
        %TwinsP-B \cite{twins} & 74 & 977  & 47.1 & 48.4 \\
        Twins-B \cite{twins} & 89  & 1020 & 47.7 & 48.9 \\
        Swin-S \cite{liu2021swin}  & 81 & 1038 & 47.6 & 49.5 \\
        Focal-T \cite{yang2021focal} & 85 & 1130 & 48.0 & 50.0 \\
        %Shuffle-S \cite{shuffleTR} & 81  & 1044 & 48.4 & 49.6 \\
        CrossFormer-B~\cite{wang2021crossformer} & 84 & 1079 & 49.2 & 50.1 \\        
        UniFormer-B \cite{li2022uniformer} & 80 & 1106 & 49.5 & 50.7 \\   
        CSWin-S~\cite{dong2021cswin} & 65 & 1027 & 50.4 & 51.5 \\     
        \rowcolor{Graylight} SViT-B  & 80 & 1036 & \textbf{50.7} &  \textbf{51.9} \\
        \midrule
         {Swin-B}~\cite{liu2021swin}  \cite{liu2021swin}  &  {121}&  {1188} & {48.1} & {49.7} \\
         {Focal-B}~\cite{yang2021focal} \cite{yang2021focal} &  {126} & {1354}&  {49.0} & {50.5} \\
         CSWin-B~\cite{dong2021cswin} & 109 & 1222 & 51.1 & 52.2 \\
          \rowcolor{Graylight}  STViT-L  & 125 & 1151 & \textbf{52.4} & \textbf{53.2} \\
        \bottomrule
    \end{tabular}
    }
\end{table}

\subsection{Semantic Segmentation}

\textbf{Settings.}
We conduct semantic segmentation experiments on the
ADE20K dataset\cite{zhou2019semantic}. We adopt Upernet \cite{xiao2018unified} as the segmentation heads and replace the backbones with our STViT-B. We follow the setting in Swin Transformer~\cite{liu2021swin} to train Upernet  for 160k iterations. More details are provided in the Appendix C.

\textbf{Results.} We provide the comparison results on semantic segmentation in Table~\ref{results_segmentation}.
Our STViT-B achieves  \textbf{+3.1} higher mIOU than Swin transformer with similar model size and can even achieve  \textbf{+2.6} higher mIOU than  Swin transformer with larger model size. It outperforms Uniformer by \textbf{+1.2} mIOU and  \textbf{+1.2} Multi-Scale(MS) mIOU.
The superior semantic segmentation performance further validates the effectiveness of our method.

\begin{table}[t]
	\centering
    \caption{Ablations of SViT.}\vspace{-0.3cm}
    \label{results_ablation}
    \resizebox{0.45\textwidth}{!}{
    \setlength{\tabcolsep}{8pt}{
    \begin{tabular}{l|c|c|c|c|c|c}
        \hline
       \multirow{2}{*}{Model} & \multicolumn{3}{c|}{ImageNet-1K} & \multicolumn{2}{c|}{COCO} & \multicolumn{1}{c}{ADE20K} \\
        & \multicolumn{1}{c}{\#Param} & \multicolumn{1}{c}{FLOPs} & \multicolumn{1}{c|}{Acc.} & \multicolumn{1}{c}{AP$^b$} & \multicolumn{1}{c|}{AP$^m$} & \multicolumn{1}{c}{mIoU} \\
         \hline
        DeiT-S & 22M & 4.6G &  79.9\% & - & - & -  \\
        Swin-T & 28M & 4.5G &  81.3\% & 42.2 & 39.1 &44.5  \\
        STA-4stage  & 24.1M & 3.97G  & 82.3\% & 43.6 &40.1 &46.0  \\
         \hline
        + Conv Stem & 24.2M & 4.26G  & 82.6\% & 44.6&41.1&46.9 \\
        + Projection & 25.2M & 4.29G  & 82.9\% &44.6&41.2&47.0 \\
        \hline
        + CPE & 25.3M & 4.30G  & 83.3\% & 46.8 & 42.5 & 47.7  \\
        w/o CPE, + APE & 24.2M & 4.26G  & 83.1\% & 45.2 &41.5 & 47.3 \\
        w/o CPE, + RPE & 25.3M & 4.29G  & 83.2\% & 45.6 &41.8&47.5\\
         \hline
        + ConvFFN & 25.4M & 4.37G & 83.6\% & 47.6 &43.1&48.6 \\
        w/o shortcut &  25.4M & 4.37G  & 83.4\% & 47.5 &43.0&48.4\\
         \hline
    \end{tabular}
   }}\vspace{-0.3cm}
\end{table}

\subsection{Ablation Study}

We conduct ablation studies to examine the role of each component of STViT. In Table \ref{results_ablation}, we start with a hierarchical 4-stage network with STA and then progressively
add the rest modules of STViT, i.e., the convolution stem, the $1\times1$ convolution projection, the CPE and the ConvFFN. The performance gradually increases, implying the effectiveness of each component. %More  discussions about the STT block are detailed as follows.

\textbf{Super Token Attention.}
As shown in Table \ref{results_ablation},
we firstly develop a simple hierarchical ViT backbone with STA, which has 4 stages like STViT-S. With no convolutional modules, it outperforms DeiT-S and Swin-T by 2.4\% and 1.0\%, respectively.
Note that it does not use position embedding while DeiT uses absolute PE, and Swin uses relative PE. The superior performance validate the effectiveness of STA.

Based on the STA-4stage backbone, we compare different variants of STA and report the results in Table~\ref{results_ablation_sta}. Specifically, we vary the grid sizes (the initial sizes of super tokens) and the sampling iterations. Note that when the grid size equals to 1, we ignore the super tokens sampling  process and  view the tokens themselves  as super tokens.

%\begin{table}[tp]
\begin{wraptable}{r}{4.3cm}
	\centering
    \caption{Ablations of STA.}\vspace{-0.3cm}
    \label{results_ablation_sta}
    \resizebox{0.25\textwidth}{!}{
    \begin{tabular}{ccccc}
       \toprule
       % Grid size & 8,1,1,1 & 8,4,1,1 &8,4,2,1 & 8,4,1,1 &4,2,1,1 &8,4,1,1 &8,4,1,1 & 8,4,1,1  \\
%        Iteration & 1 & 1 & 1 & 1 & 1 &  0 & 1 & 2 \\
%        FLOPs & 4.97G & 3.97G & 3.27G & 3.97G & 4.15G & 3.93G & 3.97G & 4.00G \\
%        Top-1 & 81.7\% &  82.3\% &  81.3\% & 82.3\%  &  82.1\% &  81.9\%  & 82.3\% & 82.1\% \\
      %  
        Grid size & Iteration & FLOPs  & Top-1  \\	
        \midrule
        8,1,1,1 & 1 & 4.97G & 81.7\% \\
        8,4,1,1 & 1 & 3.97G & 82.3\%  \\
        8,4,2,1 & 1 & 3.27G & 81.3\% \\
        \midrule
        8,4,1,1 & 1 & 3.97G & 82.3\% \\
        4,2,1,1 & 1 & 4.15G & 82.1\% \\
       % 2,1,1,1 & 1 & 5.28G & 81.5\% \\
        \midrule
        8,4,1,1 & 0 & 3.93G & 81.9\% \\
        8,4,1,1 & 1 & 3.97G & 82.3\% \\
        8,4,1,1 & 2 & 4.00G & 82.1\% \\
        %\bottomrule
        \hline
    \end{tabular}
   }
%\end{table}
\end{wraptable}

Table~\ref{results_ablation_sta} shows
that applying super token sampling works well in the first two stags but would hurt performance in the third stage. It is reasonable since super tokens are developed to address local redundancy in the shallow layers.
We also observe that larger grid sizes yield better performance. This may be attributed to the fact that larger super tokens can cover more tokens and capture global representations better .

As shown in Table~\ref{results_ablation_sta}, using a single iteration achieves the best performance. When the number of iterations is zero, the super tokens are downsampled features by average pooling and may cover different semantic regions, resulting in performance drop. Using two sampling iterations achieves a slightly worse performance with 0.03G more FLOPs. This indicates that a single iteration is enough for STA.

\textbf{Convolution Position Embedding.}
The results  in Table \ref{results_ablation} show that adding position information is vital for STViT. All the PEs can improve the performance, e.g., APE by 0.2\%,  RPE by 0.3\% and CPE by 0.4\%. STViT adopts CPE since it is more flexible for arbitrary resolutions.

\textbf{Convolution FFN.} We use ConvFFN to enhance local features and the performance gain by ConvFFN validates its effectiveness. The shortcut connection in ConvFFN is also helpful for the final performance.

\section{Conclusion}
\label{sec:concl}

We present a super token vision transformer (STViT) to learn efficient and effective global representations in the shallow layers. We adapt the design of superpixels into the token space and introduce super tokens that aggregate similar tokens together. Super tokens have distinct patterns, thus reducing local redundancy. The proposed super token attention decomposes vanilla global attention into multiplications of a sparse association map and a low-dimensional attention, leading to high efficiency in capturing global dependencies.
Extensive experiments on various vision tasks, including image classification, object detection, instance segmentation and semantic segmentation, demonstrate the effectiveness and superiority of the proposed STViT backbone.

\section*{Acknowledgment}

This work is partially funded by National Key Research and Development Program of China (Grant No. 2020AAA0140000), National Natural Science Foundation of China (Grant No. 62006228, U21B2045), and Youth Innovation Promotion Association CAS (Grant No. 2022132).

%%%%%%%%% REFERENCES
{\small
\bibliographystyle{ieee_fullname}
\bibliography{egbib}
}

\begin{appendices}

In this supplementary material, we first describe the implementation details of Super Token Attention. Then we provide the architecture details of STViT, following by the experimental settings. We also provide ablation study under isotropic settings. At last, we briefly discuss about the limitations and broader impacts of our study.

\section{Algorithm of Super Token Attention}

The proposed Super Token Attention (STA) consists of three processes, i.e., Super Token Sampling (STS), Multi-Head Self-Attention (MHSA), and Token Upsampling (TU). In particular, STS can be further decomposed into two iterative steps, reformulated as:
\begin{align}\label{eq:asso}
  Q^t &= {\rm Softmax} (\frac{X{S^{t-1}}^{\rm T}}{\sqrt{d}}), \\
  S^t &= ({{\hat{Q}}^{t}})^{\rm T}X,
\end{align}
where $\hat{Q}^t$ is the column-normalized $Q^t$. For $\upsilon$ iterations,  the complexity of the above steps is $2 \upsilon mNC$, which is relatively high. To reduce the complexity, we set $\upsilon$ to $1$ and compute the association $Q^t$ in a sparse manner. For each token, only its $3\times 3$ surrounding super tokens are used to compute $Q$ (we omit the index for clarity).
We use the Unfold and Fold functions to extract and combine the corresponding $3\times 3$ super tokens, respectively.

The computation details of STA are illustrated in Algo. \ref{alg:code}.
Given the input tokens $X\in \mathbb{R}^{C\times H\times W}$, we first generate the initial super tokens $S\in \mathbb{R}^{C\times p\times q}$ by average pooling, where $p=\frac{H}{h}$, $q=\frac{W}{w}$, and $h\times w$ is the grid size.
We then extract the $3\times 3$ super tokens $\bar{S}\in \mathbb{R}^{pq\times C\times 9}$ corresponding to each token via the Unfold function and compute the association $Q\in \mathbb{R}^{(pq\times hw)\times 9}$. We update $\bar{S}$ and combine the surrounding tokens to $S$ via the Fold function. $S$ is divided by the combined sum of $Q$ for normalization.
Since the iteration num is set to $1$, we perform the multi-head self-attention on $S$.
Finally, we perform the sparse multiplication of $Q$ and $S$ via the Unfold function like above to map super tokens back to the token space.

\begin{spacing}{1}
\begin{algorithm}[H]
\small
\caption{\small Pseudocode of Super Token Attention}
\label{alg:code}
\begin{lstlisting}[language=python]
# input: (B, C, H, W), output: (B, C, H, W)
# grid_size: (h, w)
# super token number: m=p*q, where p=H//h, q=W//w
# number of iterations: n_iter, default as 1
# scale: C**-0.5, eps: 1e-12

import torch.nn.functional as F
from einops import rearrange

def MultiHeadSelfAttention(x): return x

# rearrange the tokens
tokens = rearrange(x, "bc(yh)(xw)->b(yx)(hw)c", y=p, x=q)

# compute the initial super tokens
stokens = F.adaptive_avg_pool2d(x, (p, q))

# compute the associations iteratively
for idx in range(n_iter):
    # extract the 9 surrounding super tokens
    stokens = F.unfold(stokens, kernel_size=3)
    stokens = stokens.transpose(1, 2).reshape(B, p*q, C, 9)

    # compute sparse associations (B, p*q, h*w, 9)
    association = tokens @ stokens * scale
    association = association.softmax(-1)

    # prepare for association normalization
    association_sum = association.sum(2).transpose(1, 2).reshape(B, 9, p*q)
    association_sum = F.fold(association_sum, output_size=(p, q), kernel_size=3)

    # compute super tokens
    stokens = tokens.transpose(-1, -2) @ association
    stokens = stokens.permute(0, 2, 3, 1).reshape(B, C*9, p*q)
    stokens = F.fold(stokens, output_size=(p, q), kernel_size=3)
    stokens = stokens/(association_sum + eps)

# MHSA for super tokens
stokens = MultiHeadSelfAttention(stokens)

# map super tokens back to tokens
stokens = F.unfold(stokens, kernel_size=3)
stokens = stokens.transpose(1, 2).reshape(B, p*q, C, 9)
tokens = stokens @ association.transpose(-1, -2)
output = rearrange(tokens, "b(yx)c(hw)->bc(yh)(xw)", y=p, x=q)


\end{lstlisting}
\end{algorithm}
\end{spacing}

\begin{table*}[t]
 \centering
 \caption{Architectures for ImageNet classification with resolution $224\times 224$.}
 \label{tab:arch}
\begin{tabular}{ccccc}
\toprule
Output Size & Layer Name & STViT-S & STViT-B & STViT-L \\
\midrule
$56\times 56$ & \begin{tabular}{c} Conv Stem  \end{tabular} & $\begin{array}{c} 3\times3, 32, \text{stride}~2 \\ 3\times3, 32 \\ 3\times3, 64, \text{stride}~2 \\ 3\times3, 64   \end{array}$ & $\begin{array}{c} 3\times3, 48, \text{stride}~2 \\ 3\times3, 48 \\ 3\times3, 96, \text{stride}~2 \\ 3\times3, 96   \end{array}$ & $\begin{array}{c} 3\times3, 48, \text{stride}~2 \\ 3\times3, 48 \\ 3\times3, 96, \text{stride}~2 \\ 3\times3, 96   \end{array}$\\
\midrule
\begin{tabular}{c} Stage 1 \end{tabular} & \begin{tabular}{c}CPE \\STA\\ ConvFFN\end{tabular} & $\begin{bmatrix}\setlength{\arraycolsep}{1pt} \begin{array}{c}
		3\times3, 64\\ \text{grid}~8, \text{heads}~1\\ 3\times3, 64
		\end{array} \end{bmatrix} \times 3$ &  $\begin{bmatrix}\setlength{\arraycolsep}{1pt} \begin{array}{c}
		3\times3, 96\\ \text{grid}~8, \text{heads}~2\\ 3\times3, 96
		\end{array} \end{bmatrix} \times 4$   &  $\begin{bmatrix}\setlength{\arraycolsep}{1pt} \begin{array}{c}
		3\times3, 96\\ \text{grid}~8, \text{heads}~2\\ 3\times3, 96
		\end{array} \end{bmatrix} \times 4$ \\
\midrule
$28\times 28$ & Patch Merging & $3\times3, 128, \text{stride}~2$ & $3\times3, 192, \text{stride}~2$ &  $3\times3, 192, \text{stride}~2$ \\
\midrule
\begin{tabular}{c} Stage 2 \end{tabular} & \begin{tabular}{c}CPE \\STA\\ ConvFFN\end{tabular} & $\begin{bmatrix}\setlength{\arraycolsep}{1pt} \begin{array}{c}
		3\times3, 128\\ \text{grid}~4, \text{heads}~2\\ 3\times3, 128
		\end{array} \end{bmatrix} \times 5$ & $\begin{bmatrix}\setlength{\arraycolsep}{1pt} \begin{array}{c}
		3\times3, 192\\ \text{grid}~4, \text{heads}~3\\ 3\times3, 192
		\end{array} \end{bmatrix} \times 6$& $\begin{bmatrix}\setlength{\arraycolsep}{1pt} \begin{array}{c}
		3\times3, 192\\ \text{grid}~4, \text{heads}~3\\ 3\times3, 192
		\end{array} \end{bmatrix} \times 7$ \\
\midrule
$14\times 14$ & Patch Merging & $3\times3, 320, \text{stride}~2$ & $3\times3, 384, \text{stride}~2$ & $3\times3, 448, \text{stride}~2$ \\
\midrule
\begin{tabular}{c} Stage 3 \end{tabular} & \begin{tabular}{c}CPE \\STA\\ ConvFFN\end{tabular} & $\begin{bmatrix}\setlength{\arraycolsep}{1pt} \begin{array}{c}
		3\times3, 320\\ \text{grid}~1, \text{heads}~5\\ 3\times3, 320
		\end{array} \end{bmatrix} \times 9$ & $\begin{bmatrix}\setlength{\arraycolsep}{1pt} \begin{array}{c}
		3\times3, 384\\ \text{grid}~1, \text{heads}~6\\ 3\times3, 384
		\end{array} \end{bmatrix} \times 14$ &  $\begin{bmatrix}\setlength{\arraycolsep}{1pt} \begin{array}{c}
		3\times3, 448\\ \text{grid}~1, \text{heads}~7\\ 3\times3, 448
		\end{array} \end{bmatrix} \times 19$\\
\midrule
$7\times 7$ & Patch Merging & $3\times3, 512, \text{stride}~2$ & $3\times3, 512, \text{stride}~2$ & $3\times3, 640, \text{stride}~2$ \\
\midrule
\begin{tabular}{c} Stage 4 \end{tabular} & \begin{tabular}{c}CPE \\STA\\ ConvFFN\end{tabular} & $\begin{bmatrix}\setlength{\arraycolsep}{1pt} \begin{array}{c}
		3\times3, 512\\ \text{grid}~1, \text{heads}~8\\ 3\times3, 512
		\end{array} \end{bmatrix} \times 3$ &$\begin{bmatrix}\setlength{\arraycolsep}{1pt} \begin{array}{c}
		3\times3, 512\\ \text{grid}~1, \text{heads}~8\\ 3\times3, 512
		\end{array} \end{bmatrix} \times 6$ & $\begin{bmatrix}\setlength{\arraycolsep}{1pt} \begin{array}{c}
		3\times3, 640\\ \text{grid}~1, \text{heads}~10\\ 3\times3, 640
		\end{array} \end{bmatrix} \times 8$ \\
\midrule
$7\times 7$ & Projection & \multicolumn{3}{c}{$1\times1, 1024$} \\
\midrule
$1\times 1$ & Classifier & \multicolumn{3}{c}{Fully Connected Layer, 1000} \\
\midrule
\multicolumn{2}{c}{\# Params} & $25$ M & $52$ M & $95$ M  \\
\midrule
\multicolumn{2}{c}{\# FLOPs} & $4.4$ G & $9.9$ G & $15.6$ G  \\
\bottomrule
\end{tabular}
\end{table*}

\section{Architecture Details}

The architecture details are illustrated in Table~\ref{tab:arch}.
For the convolution stem, we adopt four $3\times 3$ convolutions to embed the input image into tokens.
GELU and batch normalization are used after each convolution.
$3\times 3$ convolutions with stride 2 are used between stages to reduce the feature resolution.
$3\times 3$  depth-wise convolutions are adopted in CPE and ConvFFN to enhance the capacity of local modeling.
The projection layer is composed of a $1\times 1$ convolution, a batch-normalization layer, and a Swish activation.
Global average pooling is used after the projection layer, following by the fully-connected classifier.

\section{Experimental Settings}

\paragraph{ImageNet Image Classification.}
We adopt the training strategy proposed in DeiT~\cite{touvron2021training}.
In particular, our models are trained from scratch for 300 epochs with the input resolution of $224\times 224$.
The AdamW optimizer is applied with a cosine decay learning rate scheduler and 5 epochs of linear warm-up.
The initial learning rate,  weight decay, and  batch-size are set to  0.001, 0.05, and 1024, respectively.
We apply the same data augmentation and regularization used  in  DeiT~\cite{touvron2021training}. Specifically, the augmentation settings are RandAugment \cite{cubuk2020randaugment} (randm9-mstd0.5-inc1) , Mixup \cite{zhang2018mixup} (prob = 0.8), CutMix \cite{yun2019cutmix} (prob = 1.0), Random Erasing \cite{zhong2020random} (prob = 0.25).
We do not use Exponential Moving Average (EMA) \cite{polyak1992acceleration}.
The  maximum rates of increasing stochastic depth \cite{huang2016deep} are set to 0.1, 0.4, 0.6 for STViT-S, STViT-B, STViT-L, respectively.
For $384\times 384$ input resolution, the models are fine-tuned for 30 epochs with learning rate of 1e-5,  weight decay of 1e-8 and bach-size of 512.

\paragraph{COCO Object Detection and Instance Segmentation.}
We apply Mask-RCNN~\cite{he2017mask} and Cascaded Mask R-CNN~\cite{cai2018cascade} as the detection heads based on MMDetection \cite{chen2019mmdetection}.
The models are trained under two common settings: ``1 $\times$" (12 training epochs) and ``3 $\times$ +MS" (36 training epochs with multi-scale training).
For the ``1 $\times$" setting, images are resized to the shorter side of 800 pixels while the longer side is within 1333 pixels.
For the ``3 $\times$ +MS", we apply the multi-scale training strategy to randomly resize the shorter side between 480 to 800 pixels.
We apply the AdamW optimizer with the initial learning rate of 1e-4 and weight decay of 0.05.
We set the stochastic depth rates to 0.1 and 0.3 for our small and base models, respectively.

\paragraph{ADE20K Semantic Segmentation.}
We apply our model as the backbone network and Upernet \cite{xiao2018unified} as the segmentation head  based on MMSegmentation \cite{contributors2020mmsegmentation}.
We follow the setting used in  Swin Transformer~\cite{liu2021swin}  and train the model for  160k iterations with the input resolution of $512\times 512$.
We set the stochastic depth rate to 0.3 for our STViT-B backbone.

%\section{Visualizations}

%\section{More Experimental Results}

%\section{High-Resolution Image Classification}

%\section{Redundancy Analysis}

% 定性比较；
% 定量比较

%\section{Visualizations} 观察不太理想，不放了

%\subsection{Attention Maps}

%\subsection{CAM}

\section{Isotropic comparison}

We have conducted comparisons under isotropic settings.
For fair comparison, as shown in Tab.~\ref{results_compare_swin}, we compare STViT against Swin with three variants according to the number of blocks, channels, and using relative position encoding (RPE) or not.
STViT outperforms Swin under similar settings, implying that STA itself can bring large performance gain.

\begin{table}[tp]
	\centering
    \caption{Comparing isotropic STViT and Swin without Conv Stem, Projection, CPE and ConvFFN.  Swin$^\dag$ uses global SA in 3rd stage. }\vspace{-0.3cm}
    \label{results_compare_swin}
    \resizebox{0.48\textwidth}{!}{
    \begin{tabular}{ccc|lccc}
        \hline
      Blocks & Channels & RPE & Model & \#Params & FLOPs & Acc. \\
	   \hline
      $[96,192,384,768]$ & $[2,2,6,2]$ & $\surd$  & Swin & 28M & 4.5G & 81.3\% \\
      $[96,192,384,768]$ &$[2,2,6,2]$ & $\surd$  & STViT & 30M & 4.3G & \textbf{82.3\%} \\
      \hline
      $[64,128,320,512]$ & $[3,5,9,3]$ & $\surd$  & Swin & 23M & 4.2G & 81.4\% \\
      $[64,128,320,512]$ & $[3,5,9,3]$ & $\surd$  & Swin$^\dag$ & 23M & 4.3G & 81.7\% \\
      $[64,128,320,512]$ & $[3,5,9,3]$ & $\surd$  & STViT & 24M & 4.0G & \textbf{82.6\%} \\
      \hline
      $[64,128,320,512]$ & $[3,5,9,3]$ & $\times$  & Swin & 23M & 4.2G & 80.2\% \\
      $[64,128,320,512]$ & $[3,5,9,3]$ & $\times$  & STViT & 24M & 4.0G & \textbf{82.3\%} \\
      \hline
    \end{tabular}
    }\vspace{-0.5cm}
\end{table}

\section{Limitations and Broader Impacts}

One possible limitation of the proposed STViT is that it adopts some operations that are not computational efficient for GPU, such as the Fold and Unfold operations in STA and the depth-wise convolutions in CPE and ConvFFN.
This makes our models not among the fastest under similar settings of FLOPs.
Besides, due to computational constraint, it is not trained on large scale datasets ( e.g., ImageNet-21K), which will be explored in the future.

The proposed STViT is a general vision transformer that can be applied on different vision tasks, e.g., image classification, object detection and semantic segmentation. It has no direct negative social impact. Possible malicious uses of STViT as a general-purpose backbone are beyond the scope of our study to discuss.

\end{appendices}

\end{document}